\documentclass[11pt,a4paper]{article}

\usepackage[T1]{fontenc}
\usepackage[utf8]{inputenc}
\usepackage{lmodern}
\usepackage[a4paper,margin=25mm]{geometry}
\usepackage{microtype}
\usepackage{textcomp}
\usepackage{graphicx}
\usepackage{amsmath,amssymb}
\usepackage{float}
\usepackage[numbers,sort&compress]{natbib}
\usepackage{xcolor}
\definecolor{linkblue}{RGB}{0,70,140}
\usepackage[
  unicode=true,
  colorlinks=true,
  linkcolor=linkblue,
  citecolor=linkblue,
  urlcolor=linkblue,
  bookmarksopen=true
]{hyperref}

\title{High-fidelity Digital Twin Data Models by Randomized Dynamic Mode Decomposition and Deep Learning with Applications in Fluid Dynamics}
\author{Diana A. Bistrian\\
\small University Politehnica Timisoara, Department of Electrical Engineering and Industrial Informatics, Romania\\
\small \href{mailto:diana.bistrian@upt.ro}{diana.bistrian@upt.ro}}
\date{}

\hypersetup{
  pdftitle={High-fidelity Digital Twin Data Models by Randomized Dynamic Mode Decomposition and Deep Learning with Applications in Fluid Dynamics},
  pdfauthor={Diana A. Bistrian},
  pdfsubject={Digital twin data models, randomized dynamic mode decomposition, and deep learning},
  pdfkeywords={digital twin data model; randomized dynamic mode decomposition; shock wave phenomena; deep learning}
}

\begin{document}
\maketitle

\begin{center}
\small
Originally published in \emph{Modelling} \textbf{3} (2022), no.~3, 314--332.\\
Received 9 June 2022; accepted 19 July 2022; published 21 July 2022.\\
\href{https://doi.org/10.3390/modelling3030020}{https://doi.org/10.3390/modelling3030020}
\end{center}

\begin{abstract}
The purpose of this paper is the identification of high-fidelity digital twin data models from numerical code outputs by non-intrusive techniques (i.e. not requiring Galerkin projection of the governing equations onto the reduced modes basis). In this paper we define the concept of digital twin data model (DTM) as a model of reduced complexity that has the main feature to mirror the original process behavior. The significant advantage of a DTM is to map the dynamics with high accuracy and reduced costs in CPU time and hardware, to timescales difficult to explore because of the complexity of the dynamics over time.
This paper introduces a new framework for creating efficient digital twin data models by combining two state-of-the-art tools: randomized dynamic mode decomposition and deep learning. We show that the outputs are consistent with the original source data with the advantage of a reduced complexity.
The DTMs are investigated in the numerical simulation of three shock wave phenomena with increasing complexity. We perform a thorough assessment of the performance of the new digital twin data models in terms of numerical accuracy and computational efficiency.
\end{abstract}

\noindent\textbf{Keywords:} digital twin data model; randomized dynamic mode decomposition; shock wave phenomena; deep learning
\bigskip

\section{Introduction}\label{secINTRO}

Data-driven algorithms for analyzing complex systems are of growing interest.
Among fluid dynamics researchers, many efforts were directed in recent years to identification of a reliable approximation of the complex flow dynamics by models of low complexity, i.e. reduced order models (ROM). In order to reveal underlying physical processes in an appropriate way, a dynamic analysis should be made. Modal decomposition techniques are superior to other techniques, because they are able to link physical behaviour to a spatial pattern.

Among several  modal decomposition methods, Proper Orthogonal Decomposition (POD) \cite{cao2007, dimistefanav2015, Du2013, Stefanescu2013, Wang2012, Mura2021} and Dynamic Mode Decomposition (DMD) \cite{Mezic2005, Rowley2010,  Bagheri2013, Schmid2012, Frederich2011, Balajewicz2013, bisnavon2016} have been widely applied to study the physics of the dynamics of the flows in different applications.

The POD and its variants are also known as Karhunen-Loeve expansions in feature selection and signal processing, empirical
orthogonal functions in atmospheric science or principal component analysis in statistics. The strong point of POD is that it can be applied to non-linear partial differential equations, especially for smooth systems in which the energetics can be hierarchically ranged and characterized by
the first few modes. The applicability of POD to complex systems is limited mainly due to errors associated with the truncation of the POD modes
\cite{Chen2011, Dimitriu2017, Bistrian2014}.

Being rooted in the Koopman mode theory \cite{Koopm1931}, a recent decomposition technique, namely Dynamic Mode Decomposition (DMD) \cite{Mezic2005, Schmid2008, Rowley2010}, has the significant advantage of linking a spatial structure (coherent structure) to a single oscillating frequency and growing/decay rate. Therefore, DMD is promising, especially for hydrodynamic research like flow field analysis.

Since the first application of the Koopman theory for the purposes of reduced order modelling by Igor Mezi\'{c} \cite{Mezic2005}, a considerable amount of work has focused on understanding and improving the method of dynamic mode decomposition and several DMD procedures have been released: optimized DMD
\cite{Chen2012}, exact DMD \cite{TU2014}, sparsity promoting DMD \cite{Jovanovic2012}, multi-resolution DMD \cite{kutz2015, Kutz2016}, extended DMD  \citep{Williams2015}, recursive DMD \citep{noack2015, noack2016}, DMD with control  \cite{Proctor2016}, randomized DMD \cite{randomized2017, Bistrian2018}, dynamic mode decomposition with core sketch \cite{OmerSan2022}.
Under assumptions on the underlying dynamics, Mezi\'{c} \cite{Mezic2022} provided the first result on the convergence rate under sample size increase in the case of finite-section approximation and introduced a discussion on the choice of observables in the
context of finite-section approximations.

A comparison of DMD vs. POD for model reduction was illustrated in our previous paper \cite{Bistrian2014}, for the study of shallow water equations model.  A procedure of coupling POD and DMD for nonlinear model order reduction was introduced in \cite{Alla2016}.

Modelling fluid dynamics data is even more difficult when we handle with discrete or so called \textit{non-intrusive data}, especially when there is no mathematical model associated with the data.

In this paper we introduce the concept of digital twin data model (DTM) as a model of reduced complexity that has the main feature to mirror the original process behavior. The significant advantage of a DTM is to map the dynamics with high accuracy and reduced costs in CPU time and hardware, even to timescales difficult to explore because of the rapidly changing dynamics over time.
This paper introduces a new framework for creating efficient digital twin data models from non-intrusive data by combining two state-of-the-art tools: randomized dynamic mode decomposition introduced in \cite{randomized2017} and deep learning artificial intelligence. We show that the outputs are consistent with the original source data with the advantage of a reduced complexity.
The DTMs are investigated in the numerical simulation of  three shock wave phenomena with increasing complexity. We perform a thorough assessment of the performance of the new digital twin data models in terms of numerical accuracy and computational efficiency.

The remainder of the article is organized as follows. In Section 2 the test problem consisting of the nonlinear viscous Burgers equation
model is presented. In Section 3 we recall the principles governing the dynamic mode decomposition and we provide the description of the randomized dynamic mode decomposition algorithm in Section 4. In Section 5 we outline the technique of fast digital twin data model identification using deep learning Nonlinear Autoregressive Estimators. Section 6 presents the numerical results together with a computational efficiency study. Summary and conclusions are drawn in the final section.

\section{Shock Wave Phenomena: Full-order Model of Nonlinear Viscous Burgers Equation}

We consider that the experimental data are provided by the simulation of the nonlinear viscous Burgers equation model \cite{Burgers1948})
\begin{equation}\label{Burgers}
\left\{ {\begin{array}{*{20}{l}}
{\frac{\partial }{{\partial t}}u\left( {x,t} \right) + \frac{\partial }{{\partial x}}\left( {\frac{{u{{\left( {x,t} \right)}^2}}}{2}} \right) = \nu \frac{{{\partial ^2}}}{{\partial {x^2}}}u\left( {x,t} \right),\quad x \in \left[ {0,L} \right],\;t \in \left[ {0,T} \right],}\\
{u\left( {x,0} \right) = {u_0}\left( x \right),}
\end{array}} \right.
\end{equation}
where $u\left( {x,t} \right)$ is the unknown function of time $t$, $\nu  = 1/{\mathop{\rm Re}\nolimits} $ is the viscosity term and $Re$ is the Reynolds number. We consider the discontinuous initial condition of the form
\begin{equation}\label{Inicond}
{u_0}\left( x \right) = \left\{ \begin{array}{l}
{u_L},\quad x \le 0,\\
{u_R},\quad x > 0.
\end{array} \right.
\end{equation}

This setting yields a shock wave phenomenon.
The the initial value problem (\ref{Burgers})-(\ref{Inicond}) is solved using a finite difference discretization of the conservative form of the
equation (\ref{Burgers}), and then carrying out a parabolic integration scheme \cite{Cuesta2009}. The constants used for the test model are
\[L = 2,\quad T = 3,\quad {u_L} = 0.1,\quad {u_R} = 0.5.   \]

The training data comprises of

${N_t}+1=301$ total number of snapshots taken in time at regularly spaced time intervals $\Delta t = 0.01$,

${N_x}=101$ number of spatial measurements per time snapshot.

The nonlinear evolution governed by Burgers equation can be obtained also by
the Cole-Hopf transformation \cite{KutzHopf}.

The Cole-Hopf transformation is defined by
\begin{equation}\label{cole}
u =  - 2\nu \frac{1}{\varphi }\frac{{\partial \varphi }}{{\partial x}}.
\end{equation}

Through an analytical handling we find that
\begin{equation}\label{e1}
\frac{{\partial u}}{{\partial t}} = \frac{{2\nu }}{{{\varphi ^2}}}\left( {\frac{{\partial \varphi }}{{\partial t}}\frac{{\partial \varphi }}{{\partial x}} - \varphi \frac{{{\partial ^2}\varphi }}{{\partial x\partial t}}} \right),\quad u\frac{{\partial u}}{{\partial x}} = \frac{{4{\nu ^2}}}{{{\varphi ^3}}}\frac{{\partial \varphi }}{{\partial x}}\left( {\varphi \frac{{{\partial ^2}\varphi }}{{\partial {x^2}}} - \frac{{\partial \varphi }}{{\partial x}}\frac{{\partial \varphi }}{{\partial x}}} \right),
\end{equation}
\begin{equation}\label{e2}
\nu \frac{{{\partial ^2}u}}{{\partial {x^2}}} =  - \frac{{2{\nu ^2}}}{{{\varphi ^3}}}\left( {2{{\left( {\frac{{\partial \varphi }}{{\partial x}}} \right)}^3} - 3\varphi \frac{{{\partial ^2}\varphi }}{{\partial {x^2}}}\frac{{\partial \varphi }}{{\partial x}} + {\varphi ^2}\frac{{{\partial ^3}\varphi }}{{\partial {x^3}}}} \right).
\end{equation}

Substituting these expressions into (\ref{Burgers}) it follows that
\begin{equation}\label{e3}
\frac{{\partial \varphi }}{{\partial x}}\left( {\frac{{\partial \varphi }}{{\partial t}} - \nu \frac{{{\partial ^2}\varphi }}{{\partial {x^2}}}} \right) = \varphi \left( {\frac{{{\partial ^2}\varphi }}{{\partial x\partial t}} - \nu \frac{{{\partial ^3}\varphi }}{{\partial {x^3}}}} \right) = \varphi \frac{\partial }{{\partial x}}\left( {\frac{{\partial \varphi }}{{\partial t}} - \nu \frac{{{\partial ^2}\varphi }}{{\partial {x^2}}}} \right).
\end{equation}

Relation (\ref{e3}) indicates that if $\varphi $ solves the heat equation, then $u\left( {x,t} \right)$ given by the Cole-Hopf transformation (\ref{cole}) solves the viscid Burgers equation (\ref{Burgers}). Thus we have reduced the viscid Burgers equation (\ref{Burgers}) to the following one
\begin{equation}\label{heat}
\left\{ \begin{array}{l}
\frac{{\partial \varphi }}{{\partial t}} - \nu \frac{{{\partial ^2}\varphi }}{{\partial {x^2}}} = 0,\quad x \in R,t > 0,\nu  > 0,\\
\varphi \left( {x,0} \right) = {\varphi _0}\left( x \right) = {e^{ - \int_0^x {\frac{{{u_0}(\xi )}}{{2\nu }}d\xi } }},x \in \mathbb{R}.
\end{array} \right.
\end{equation}

Taking the Fourier transform with respect to $x$ for both heat equation and the initial condition (\ref{heat}) we obtain the analytic solution
\begin{equation}\label{solphi}
\varphi \left( {x,t} \right) = \frac{1}{{2\sqrt {\pi \nu t} }}\int\limits_{ - \infty }^\infty  {{\varphi _0}\left( \xi  \right)} \,{e^{ - \frac{{{{(x - \xi )}^2}}}{{4\nu t}}}}d\xi .
\end{equation}

From the Cole-Hopf transformation (\ref{cole}) we obtain the analytic solution to the problem (\ref{Burgers}) in the following form
\begin{equation}\label{exactsolu}
u\left( {x,t} \right) = \frac{{\int_{ - \infty }^\infty  {\frac{{x - \xi }}{t}{\varphi _0}\left( \xi  \right){e^{ - \frac{{{{\left( {x - \xi } \right)}^2}}}{{4\nu t}}}}d\xi } }}{{\int_{ - \infty }^\infty  {{\varphi _0}\left( \xi  \right){e^{ - \frac{{{{\left( {x - \xi } \right)}^2}}}{{4\nu t}}}}d\xi } }}.
\end{equation}

\begin{figure}[h!]
\centering
\includegraphics[width=1\textwidth]{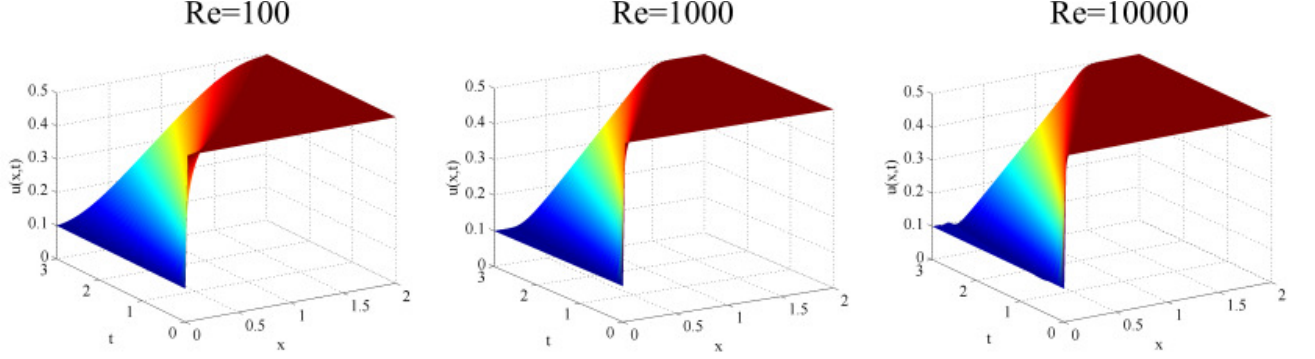}
\caption{Dynamics of shock wave phenomena as the exact solution of viscous Burgers equation model at ${\mathop{\rm Re}\nolimits}  = {10^2}$, ${\mathop{\rm Re}\nolimits}  = {10^3}$ and ${\mathop{\rm Re}\nolimits}  = {10^4}$, respectively.}\label{fig1ex}
\end{figure}

Three solution types will be discussed in the paper, corresponding to Reynolds number of ${\mathop{\rm Re}\nolimits}  = {10^2}$, ${\mathop{\rm
Re}\nolimits}  = {10^3}$ and ${\mathop{\rm Re}\nolimits}  = {10^4}$, respectively (see Figure \ref{fig1ex}). We notice that the model solution exhibits some oscillations for all three experiments. The unphysical oscillation originates due to high Reynolds numbers. We experienced that when the Reynolds number gets higher, the numerical solution exhibits more oscillations, so that the fluid dynamics become more and more complex.

We aim in this paper to identify a reduced order model of the nonlinear viscous Burgers equation model to approximate as faithful the true
solution and to create a digital twin data model of low complexity for the three shock wave phenomena, respectively. An efficient numerical technique is provided in the following sections.

\section{Reduced order modeling based on Dynamic Mode Decomposition}

\subsection{The key steps of Dynamic Mode Decomposition and snapshots collection}

We proceed by collecting data ${u_i}\left( {t,x} \right) = u\left( {{t_i},x} \right),\;{t_i} = i\Delta t,\;i = 0,...,{N_t}$, at the constant sampling
time $\Delta t$, $x$ representing the spatial coordinate.

We form a data matrix whose columns represent the individual data samples, called \textit{the snapshot matrix}:
\begin{equation}
V = \left[ {\begin{array}{*{20}{c}}
{{u_0}}&{{u_1}}&{...}&{{u_{N_t}}}
\end{array}} \right] \in {\mathbb{R}^{{N_x} \times ({N_t} + 1)}}.
\end{equation}
Each column $u_i$ is a vector with ${N_x}$ components, representing the numerical measurements. For simplicity of description, we consider here real
data ${u_i} \in {\mathbb{R}^{N_x}}$.

Following the Koopman decomposition assumption \cite{Koopm1931}, we consider that a propagator matrix $\mathcal{A}$ exists, that maps every column
vector onto the next one. The DMD algorithm constructs the best approximation of the propagator matrix $\mathcal{A}$, i.e.
\begin{equation}\label{steps1}
\left\{ {{u_0},\;{u_1} = {\cal A}{u_0},\;{u_2} = {\cal A}{u_1} = {{\cal A}^2}{u_0},.\;..,\;{u_{N_t}} = {\cal A}{u_{{N_t} - 1}} = {{\cal A}^{{N_t}}}{u_0}} \right\}.
\end{equation}

 The next computational step consists in forming two time--shifted data
matrices from the snapshot sequence. A matrix $V_0$ is formed with the first ${N_t}$ columns and the matrix $V_1$ contains the last ${N_t}$ columns of $V$:
\begin{equation}
V_0 = \left[ {\begin{array}{*{20}{c}}
{{u_0}}&{{u_1}}&{...}&{{u_{{N_t} - 1}}}
\end{array}} \right] \in {\mathbb{R}^{{N_x} \times {N_t}}},\;V_1 = \left[ {\begin{array}{*{20}{c}}
{{u_1}}&{{u_2}}&{...}&{{u_{N_t}}}
\end{array}} \right] \in {\mathbb{R}^{{N_x} \times {N_t}}}.
\end{equation}

For a sufficiently long sequence of the snapshots, we suppose that the last snapshot $u_{N_t}$ can be written as a linear combination of previous
${N_t}$ vectors, such that
\begin{equation}\label{linearcomb}
u_{N_t} = {c_0}{u_0} + {c_1}{u_1} + ... + {c_{{N_t} - 1}}{u_{{N_t} - 1}} + \mathcal{R},
\end{equation}
in which ${{\rm{c}}_i} \in {\rm{\mathbb{R},i  =  0,}}...{\rm{,{N_t}  -  1}}$  and $\mathcal{R}$ is the residual vector. We assemble the following
relations
\begin{equation}\label{kryl}
\left\{ {{u_1},{u_2},...{u_{N_t}}} \right\} = {\cal A}\left\{ {{u_0},{u_1},...{u_{{N_t} - 1}}} \right\} = \left\{ {{u_1},{u_2},...,V_0c} \right\} + {\cal R},
\end{equation}
where $c = {\left( {\begin{array}{*{20}{c}} {{c_0}}&{{c_1}}&{...}&{{c_{{N_t} - 1}}}\end{array}} \right)^T}$ is the unknown column vector.

In matrix notation form, Eq. (\ref{kryl}) reads
\begin{equation}\label{rel}
{\cal A}V_0 = V_0{\cal S} + {\cal R},\quad {\cal S} = \left( {\begin{array}{*{20}{c}}
0&{...}&0&{{c_0}}\\
1&{}&0&{{c_1}}\\
 \vdots & \vdots & \vdots & \vdots \\
0& \ldots &1&{{c_{N_t - 1}}}
\end{array}} \right),
\end{equation}
where $\mathcal{S}$ is the companion matrix.

Relation (\ref{rel}) is true when the residual
\begin{equation}\label{rez}
\mathcal{R} = {u_{N_t}} - V_0c,
\end{equation}
is minimized when $c$ is chosen such that $\mathcal{R}$ is orthogonal to $span\left\{ {{u_0},...,{u_{{N_t} - 1}}} \right\}$.

The goal of DMD algorithm is to solve the eigenvalue problem of the companion matrix $\mathcal{S}$
\begin{equation}\label{approx}
V_1 = \mathcal{A}V_0 = V_0\mathcal{S} + \mathcal{R},
\end{equation}
where $\mathcal{S}$ approximates the eigenvalues of $\mathcal{A}$ when ${\left\| \mathcal{R} \right\|_2} \to 0$, which is equivalent to solve the
minimization problem
\begin{equation}\label{minpro}
\mathop {Minimize}\limits_\mathcal{S} \;\mathcal{R} = {\left\| {V_1 - V_0\mathcal{S}} \right\|_2}.
\end{equation}

In our previous work \citep{Bistrian2014}, we estimate the solution to the minimization problem (\ref{minpro}) multiplying $V_{1}$ by the
Moore-Penrose pseudoinverse \citep{Golub1996} of $V_{0}$:
\begin{equation}
\mathcal{S}={\left( {{V_0}} \right)^ + }{V_1}.
\end{equation}
 As we previously pointed out in \citep{Bistrian2014}, the Moore-Penrose pseudoinverse approach might not be feasible when dealing with high dimensional
 data.

Following Schmid \citep{Schmid2010} who was the first to introduce the DMD as a numerical tool to compute the Koopman modes, we developed an alternate algorithm based on Singular Value Decomposition (SVD) of snapshot matrix $V_0$. This approach
is helpful especially when the matrix $V_0$ is rank deficient (${{\rm{N}}_x}{\rm{  >  }}{{\rm{N}}_t}$). In the following, we describe this technique.

We first identify a singular value decomposition of $V_{0}$:
\begin{equation}\label{svdul}
V_{0}= U\Sigma {W^H},
\end{equation}
where $U$ contains the proper orthogonal modes of $V_{0}$, $\Sigma $ is a square diagonal matrix containing the singular values of $V_{0}$ and
${W^H}$ is the conjugate transpose of $W$.

A direct consequence of solving the minimization problem (\ref{minpro}) is that decreasing the residual increases overall convergence and therefore
the eigenvalues ${\lambda _j}$ and the eigenvectors ${\phi _j}, j = 1,...,N_t$ of $\mathcal{S}$ will converge toward the eigenvalues and the
eigenvectors of the Koopman operator $\mathcal{A}$, respectively. More specifically, every column vector $u_i, i = 1,...,N_t$ can be written as a
linear combination of its predecessor:
 \begin{equation}\label{approx2}
{u_i} = \mathcal{A}{u_{i - 1}} =  \ldots  = {\mathcal{A}^{i - 1}}{u_1},\quad i = 1,...,N_t.
\end{equation}

The eigenvectors of $\mathcal{S}$ form a basis for the span of $\mathcal{A}$, therefore, we can write every column vector as a linear combination of
the eigenvectors
\begin{equation}\label{expl}
{u_i} = \sum\limits_{j = 1}^{{N_t}} {{{\cal A}^{i - 1}}{{\widetilde a}_j}{\phi _j}} \quad  \Leftrightarrow \quad {u_i} = \sum\limits_{j = 1}^{{N_t}} {{{\widetilde a}_j}\lambda _j^{i - 1}{\phi _j}} ,\quad i = 1,...,{N_t}.
\end{equation}

A straightforward interpretation of relations (\ref{expl}) brings the data snapshots at every time step $\left\{
{{{\rm{t}}_1}{\rm{,}}...{\rm{,}}{{\rm{t}}_{N_t}}} \right\}$ as a linear combination of DMD modes according to
\begin{equation}\label{allmodes}
\begin{array}{*{20}{l}}
{{V_1} = \left[ {\begin{array}{*{20}{c}}
{{u_1}}&{{u_2}}&{...}&{{u_{{N_t}}}}
\end{array}} \right] = }\\
{ = \left[ {\begin{array}{*{20}{c}}
{{\phi _1}}&{{\phi _2}}&{...}&{{\phi _{{N_t}}}}
\end{array}} \right]\left( {\begin{array}{*{20}{c}}
{{{\widetilde a}_1}}&{}&{}&{}\\
{}&{{{\widetilde a}_2}}&{}&{}\\
{}&{}& \vdots &{}\\
{}&{}&{}&{{{\widetilde a}_{{N_t}}}}
\end{array}} \right)\left( {\begin{array}{*{20}{c}}
1&{\lambda _1^1}&{\lambda _1^2}& \ldots &{\lambda _1^{{N_t} - 1}}\\
1&{\lambda _2^1}&{\lambda _2^2}& \ldots &{\lambda _2^{{N_t} - 1}}\\
1& \vdots & \vdots & \vdots & \vdots \\
 \vdots & \vdots & \vdots & \vdots & \vdots \\
1&{\lambda _{{N_t}}^1}&{\lambda _{{N_t}}^2}& \ldots &{\lambda _{{N_t}}^{{N_t} - 1}}
\end{array}} \right)}
\end{array},
\end{equation}
where the right eigenvectors of $\mathcal{S}$, ${\phi _j} \in \mathbb{C}$ are dynamic \textit{shape} (or Koopman) modes, the eigenvalues of
$\mathcal{S}$, ${\lambda _j}$ are called Ritz values \citep{Chopr2000} and coefficients ${{\widetilde a}_j} \in \mathbb{C}$ are denoted as amplitudes
or Koopman eigenfunctions. Each Ritz value ${\lambda _j} = {e^{\left( {{\sigma _j} + i{\omega _j}} \right)\Delta t}}$ is associated with the growth
rate ${\sigma _j}$ and the frequency ${\omega _j}$, where
\begin{equation}\label{sigmaomega}
{\sigma _j} = \frac{{\log \left( {\left| {{\lambda _j}} \right|} \right)}}{{\Delta t}},\quad {\omega _j} = \frac{{\arg \left( {\left| {{\lambda _j}} \right|} \right)}}{{\Delta t}}.
\end{equation}

The superposition of all Koopman modes, weighted by their amplitudes and complex frequencies, approximates the entire data sequence, but there are
also modes that have a weak contribution. Our goal is to produce the ROM involving only the most significant modes, having a strong contribution to
the data representation, which we are calling \textit{leading modes}.

Thus, the data snapshots at every time step $\left\{ {{{\rm{t}}_1}{\rm{,}}...{\rm{,}}{{\rm{t}}_{N_t}}} \right\}$ will be represented as a linear
combination of the leading DMD modes according to
\begin{equation}\label{dmdmodel1}
{u_{DMD}}\left( {{t_i},{x}} \right) = \sum\limits_{j = 1}^{N_{DMD}} {{{\widetilde a}_j}{\phi _j}\left( {x} \right)\lambda _j^{i - 1}} ,\quad i \in \left\{ {1,...,{N_t}} \right\},\quad {t_i} \in \left\{ {{{\rm{t}}_1},...,{{\rm{t}}_{N_t}}} \right\},
\end{equation}
where  ${N_{DMD}}$ represents the number of \textit{leading} DMD \textit{modes} involved in reconstruction of data snapshots.

One advantage of DMD is that each mode is associated with a pulsation, a growth rate and each mode oscillates at a single frequency, as seen from
(\ref{dmdmodel1}). Representation (\ref{dmdmodel1}) is suitable when one wants to isolate a mode with a certain frequency, or to identify a maximum or a minimum amplitude and for hydrodynamic stability analysis also.
In the seminal article \cite{Mezic2022}, Mezi\'{c} provides a characterization of Koopman modes in Banach spaces using Generalized
Laplace Analysis.

For the purpose of model order reduction, in our paper we adopt the following form

\begin{equation}\label{dmdmodel2}
{u_{DMD}}\left( {{t_i},x} \right) = \sum\limits_{j = 1}^{{N_{DMD}}} {{a_j}\left( {{t_i}} \right){\phi _j}\left( x \right)} ,\quad {t_i} \in \left\{
{{{\rm{t}}_1},...,{{\rm{t}}_{{N_t}}}} \right\},
\end{equation}
where ${\phi _j} \in \mathbb{C}$ are dynamic \textit{leading modes}  and ${a_j}\left( {{t_i}} \right) = {\widetilde a_j}\lambda _j^{i -
1},$ $i \in \left\{ {1,...,{N_t}} \right\},\;$ $j \in \left\{ {1,...,{N_{DMD}}} \right\}$
 are modal amplitudes.

Here we point out that the ${N_{DMD}}$ leading modes involved in ROM representation of data (\ref{dmdmodel2}) are not the first ${N_{DMD}}$ modes
from representation (\ref{allmodes}). The \textbf{leading modes represent a subset of DMD modes} that will be selected from all computed DMD modes via numerical algorithm presented in the next section.

\section{Offline stage: Randomized Dynamic Mode Decomposition }

The modes' selection plays a central role in model reduction and constitutes also the source of many discussions among modal decomposition
 practitioners \cite{Noak2011, Jovanovic2012, Chen2012, Tissot2014}. Several procedures for selecting the most influential
   modes in dynamic mode decomposition can be found in our previous papers \cite{Bistrian2014, aleksey2015, bisnavon2016}. We have introduced in \citep{randomized2017} the procedure of
randomization of data prior to singular value decomposition (SVD). Thus, we endow the DMD algorithm with a randomized SVD function, aiming to improve
the accuracy of the reduced order linear model and to reduce the CPU time. The major advantage of this method is that does not require an additional
selection algorithm of the DMD modes. The randomized DMD produces a reduced order subspace of Ritz values, having the same dimension as the rank of
randomized SVD function, where the leading modes live. The second advantage consists in reducing the problem dimension to avoid a computationally
expensive SVD.

The objective of the DMD-based ROM is to represent, as accurately as possible, the high fidelity solution using the dynamics given by the DMD modes.
It is then natural to seek the leading DMD modes and their temporal eigenfunctions that minimize the error
\begin{equation}\label{eror}
{E_{DMD}} = {\left\langle {{{\left\| {u\left( {x,t} \right) - {u_{DMD}}\left( {x,t} \right)} \right\|}_2}} \right\rangle _T},
\end{equation}
where ${\left\langle  \cdot  \right\rangle _T}$ is a time average operator over $\left[ {{t_1},T} \right]$ and ${\left\| {\, \cdot \,} \right\|_2}$
is the ${L_2}$-norm of ${\mathbb{R}^{{N_x}}}$. In this paper, ${\left\langle  \cdot  \right\rangle _T}$ corresponds to the arithmetic time-average on
${N_t}$ equally spaced elements of the interval $\left[ {{t_1},T} \right]$:
\begin{equation}\label{Tnorm}
{\left\langle {f\left( t \right)} \right\rangle _T} = \frac{1}{{{N_t}}}\sum\limits_{i = 1}^{{N_t}} {f\left( {{t_i}} \right)} ,\quad {t_i} \in \left\{ {{t_1},{t_2},...,{t_{{N_t}}} = T} \right\}.
\end{equation}

Determination of the optimal rank $N_{DMD}$ of the ROM then amounts to finding the solution to the following constrained optimization problem:
\begin{equation}\label{optimprob}
\left\{ {\begin{array}{*{20}{l}}
{\mathop {Find}\limits_{{N_{DMD}} \in \mathbb{N},{N_{DMD}} \ge 2} \;\;{u_{DMD}}\left( {{t_i},x} \right) = \sum\limits_{j = 1}^{{N_{DMD}}} {{a_j}\left( {{t_i}} \right){\phi _j}\left( x \right)} ,\quad {t_i} \in \left\{ {{{\rm{t}}_1},...,{{\rm{t}}_{N_t}}} \right\},}\\
{Subject\;to\;\;{N_{DMD}} = \arg \min \left\{ {{E_{DMD}}} \right\},}
\end{array}} \right.
\end{equation}
where $E_{DMD}$ is the error of the low-rank model defined by Eq. (\ref{eror}).

Generally, DMD does not produce orthogonal modes, therefore ROMs produced via DMD require a closure model consisting of additional
regularization techniques \cite{Cordier2010, San2015, WangNavon2016} especially when applying Galerkin or Petrov--Galerkin projection based
techniques.

In this paper we propose a variant of randomized dynamic mode decomposition introduced in \citep{randomized2017} augmented with deep learning artificial intelligence that confers multiple advantages to the ROM, which will be presented in the following.

The algorithm proceeds as follows:

\noindent\rule{12.5cm}{0.3pt}

\textbf{Algorithm 1: Randomized Dynamic Mode Decomposition}

\noindent\rule{12.5cm}{0.3pt}

\textbf{Initial data:} $V_0 \in {\mathbb{R}^{{N_x} \times {N_t}}}$, $V_1 \in {\mathbb{R}^{{N_x} \times {N_t}}}$, integer target rank $k \ge 2$ and $k
< {N_t}$.

\begin{enumerate}

\item[(1:)] For $k=2$ to ${N_t}-1$.

\item[(2:)] Produce the randomized singular value decomposition of rank $k$

$\left[ {U,\Sigma ,W} \right] = \textbf{$k$-RSVD}\left( V_0,k \right),$

 where $U$ contains the proper orthogonal modes of $V_0$ and $\Sigma $ contains the singular values. The RSVD function is described in continuation
 of this algorithm.

\item[(3:)] Solve the minimization problem (\ref{minpro}).

\item[4.]	Compute dynamic modes solving the eigenvalue problem $SX = X\Lambda $ and obtain dynamic modes as $\Phi  = UX$. The diagonal entries of
    $\Lambda$ represent the eigenvalues $\lambda$.

\item[5.]  Project dynamic modes onto the first snapshot to calculate the vector containing dynamic modes amplitudes $Ampl = \left( {{a_j}}
    \right)_{j = 1}^{rank\left( \Lambda  \right)}$.

\item[6.] The DMD model of rank $k$ is given by the product
\begin{equation}
V_{DMD} = \Phi  \cdot diag\left( Ampl \right) \cdot Van,
\end{equation}
where the Vandermonde matrix is
\[Van = \left( {\begin{array}{*{20}{c}}
1&{\lambda _1^1}&{\lambda _1^2}& \ldots &{\lambda _1^{N - 2}}\\
1&{\lambda _2^1}&{\lambda _2^2}& \ldots &{\lambda _2^{N - 2}}\\
1& \vdots & \vdots & \vdots & \vdots \\
 \ldots & \ldots & \ldots & \ldots & \ldots \\
1&{\lambda _k^1}&{\lambda _k^2}& \ldots &{\lambda _k^{N - 2}}
\end{array}} \right).\]

\item[7.] Solve the optimization problem (\ref{optimprob}) and obtain the optimal low rank $k$ and associated $V_{DMD}$.

\textbf{Output:} $k$, $V_{DMD}$.
\end{enumerate}

\noindent\rule{12.5cm}{0.3pt}

The following routine  is used to produce the randomized singular value decomposition.

\textbf{Algorithm 2: Randomized Singular Value Decomposition of Rank k ($k$-RSVD)}

\noindent\rule{12.5cm}{0.3pt}

\textbf{Initial data:} $V_0 \in {\mathbb{R}^{{N_x} \times {N_t}}}$, integer target rank $k \ge 2$ and $k < {N_t}$.

\begin{enumerate}

\item[(1:)] Generate random test matrix $M = rand\left( {{N_t},r} \right)$, $r = \min \left( {{N_t},2k} \right)$.

\item[(2:)] Compute sampling matrix by multiplication of snapshot matrix with random matrix $Q = V_0M$.

\item[(3:)] Orthonormalization of sampling matrix via Gram--Schmidt orthonormal method $Q \leftarrow GramSchmidt\left( Q \right)$.

\item[(4:)] Projection of snapshot matrix to smaller space $V = {Q^H}V_0$, where $H$ denotes the conjugate transpose.

\item[(5:)] Produce the economy-size singular value decomposition of low-dimensional snapshot matrix $\left[ {{T},\Sigma ,W} \right] = SVD\left(
    V \right)$.

\item[(6:)] Compute the right singular vectors $U = Q{T}$.

\textbf{Output:}  Procedure returns $U \in {\mathbb{R}^{{N_x} \times k}}$, $\Sigma  \in {\mathbb{R}^{k \times k}}$, $W \in {\mathbb{R}^{{N_t}
\times k}}$.

\end{enumerate}
\noindent\rule{12.5cm}{0.3pt}

\begin{figure}[H]
\includegraphics[width=11 cm]{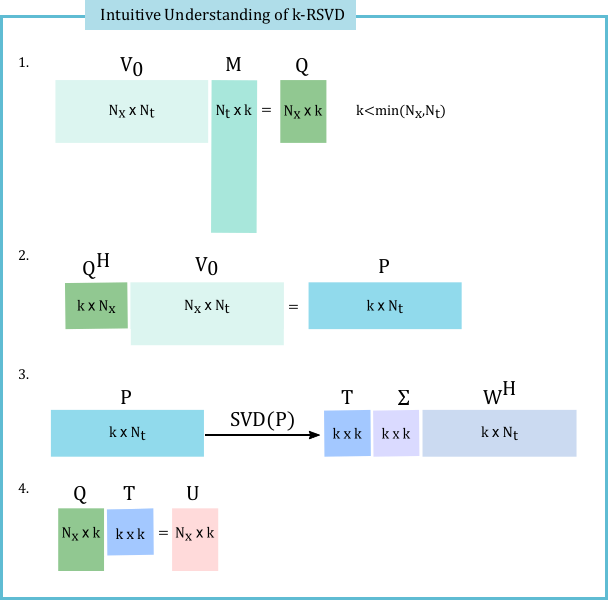}
\caption{An intuitive understanding of $k$-RSVD.\label{krsvd}}
\end{figure}

An intuitive understanding of k-RSVD is illustrated in Figure \ref{krsvd}. We avoid a computationally expensive algorithm and we reduce the problem dimension by using the randomized singular value decomposition (RSVD) technique.
\unskip

\section{Online stage: Fast Digital Twin Data Model Identification Using Deep Learning Nonlinear Autoregressive Estimators}

The algorithm previously described allows the identification of the leading dynamic modes and their associated temporal coefficients in discrete
form. The goal in this section is the identification of the reduced order digital twin data model (DTM) of the form:
\begin{equation}\label{rommodel}
u_{DTM}^{ROM}\left( {t,x} \right) = \sum\limits_{j = 1}^{{N_{DMD}}} {{{\widehat a}_j}\left( t \right){\phi _j}\left( x \right)} ,\quad t \in \left[ {0,T} \right],
\end{equation}
where ${\phi _j}$, $j = 1,...,{N_{DMD}}$ are the DMD modes and ${{{\widehat a}_j}\left( t \right)}$, $j = 1,...,{N_{DMD}}$ represent the temporal
coefficients of the DTM.

Nonlinear AutoRegressive models with eXogenous inputs (NLARX) represent a novel approach in the field of nonlinear system identification
\cite{Nare1990, Judi1995, Ljung1999}. Since the emergence of artificial neural networks as numerical tools, NLARX models have been used for various
purposes, ranging from simulation \cite{Nelles2001}, to nonlinear predictive control \cite{Liu1993} or higher order nonlinear optimization problems
\cite{Moody1991, Peng2002, Wang2018, Tieleman2012}.

We will investigate in this paper the application of NLARX models to a high-fidelity approximation of temporal coefficients of the DMD-ROM model
(\ref{rommodel}). Let $a\left( t \right)$ be  the system input represented by the DMD computed amplitudes at discrete time instances $t \in \left\{
{{{\rm{t}}_1},...,{{\rm{t}}_{{N_t}}}} \right\}$ and $\widehat a\left( t \right)$ be the output. The formulation of the NLARX model can be described
as:
\begin{equation}\label{nlarx}
\widehat a\left( t \right) = f\left[ {\widehat a\left( {t - 1} \right),...,\widehat a\left( {t - {n_a}} \right),a\left( {t - {n_k}}
\right),...,a\left( {t - {n_k} - {n_b} + 1} \right)} \right] + e\left( t \right),
\end{equation}
where the ${{n_a}}$ is the integer number of past output terms, ${{n_b}}$ is the number of past input terms used to predict the current output,
${{n_k}}$ is the pure input delay, $f$ is a nonlinear function (typically implemented by an artificial neural network) and $e\left( t \right)$
represents the modeling error. Each output of NLARX model (\ref{nlarx}) is a function of regressors that are transformations of past inputs and past
outputs. Usually this function has a linear block and a nonlinear block. The model output is the sum of the outputs of the two blocks. The NLARX
model training can be cast as a non-linear unconstrained optimization problem:
\begin{equation}\label{nlarxoptim}
\theta \left( {{n_a},{n_b},{n_k}} \right) = \arg \min \,\frac{1}{{2{N_t}}}\sum\limits_{i = 1}^{{N_t}} {{{\left\| {a\left( {{t_i}} \right) - \widehat
a\left( {{t_i}} \right)} \right\|}_2}},
\end{equation}
where the training set consists of the measured input $a\left( t \right)$, $\widehat a\left( t \right)$ is the NLARX output, ${\left\| {\, \cdot \,}
\right\|_2}$ is the ${L_2}$ norm and $\theta \left( {{n_a},{n_b},{n_k}} \right)$ represents the parameter vector of the nonlinear function $f$.

The NLARX structure can accommodate the dynamics of the system by feeding previous network outputs back into the input layer. It also enables the
user to define how many previous output and input time steps are required for a best representation of the systems dynamics. One of most important
points of an application of NLARX network is a proper selection of inputs, input delays and output delays. The task will be to modify the network
parameters $\theta \left( {{n_a},{n_b},{n_k}} \right)$ over the complete trajectory to achieve the minimal value of (\ref{nlarxoptim}).

\begin{figure}[H]
\includegraphics[width=13 cm]{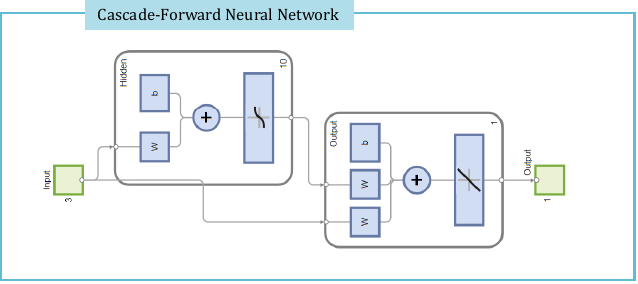}
\caption{The cascade forward neural network with 10 hidden layer sizes, used as nonlinear estimator for the NLARX models. \label{net}}
\end{figure}

We have implemented the nonlinear estimator $f$ in form of a cascade forward neural network with 10 hidden layer sizes, see Figure \ref{net}.
\unskip

In the next section, we will detail the numerical results.

\section{ Numerical Results: Computational Efficiency of the Algorithm}

In the following, we present numerical results demonstrating the computational performance of the algorithm, considering the  nonlinear
viscous Burgers equation model (\ref{Burgers}) generating three shock wave phenomena with increasing complexity. The randomization of input data has been leveraged to accelerate DMD computations.
The optimal rank of the reduced DMD model is the unique solution to the optimization problem (\ref{optimprob}). We have tested several global optimization methods like genetic algorithm combined with sequential quadratic programming (GA-SQP) \cite{Nocedal2006} and simulated annealing (SA) \cite{SA}, to solve the optimization problem (\ref{optimprob}), with similar computational efforts. A major advantage that comes from application of randomized DMD algorithm is that this leads to the optimal low rank $N_{DMD}$ and associated DMD subspace where the most influential DMD modes are identified.

The correlation coefficient defined below is used as additional metric to validate the quality of the low-rank DMD model:
\begin{equation}\label{corelation}
{C_{DMD}} = \frac{{{{\left\langle {{{\left\| {u\left( {x,t} \right) \cdot {u_{DMD}}\left( {x,t} \right)} \right\|}_2}} \right\rangle }_T}^2}}{{{{\left\langle {{{\left\| {u{{\left( {x,t} \right)}^H} \cdot u\left( {x,t} \right)} \right\|}_2}} \right\rangle }_T}{{\left\langle {{{\left\| {{u_{DMD}}{{\left( {x,t} \right)}^H} \cdot {u_{DMD}}\left( {x,t} \right)} \right\|}_2}} \right\rangle }_T}}},
\end{equation}
where $u\left( t,x \right)$ means the numerical data, ${{u_{DMD}}\left( t,x \right)}$  represent the computed solution by means of the reduced order
DMD model, $\left(  \cdot  \right)$ represents the Hermitian inner product, $H$ denotes the conjugate transpose and ${\left\langle  \cdot
\right\rangle _T}$ is the norm defined by Eq. (\ref{Tnorm}).

\begin{figure}[h!]
\centering
\includegraphics[width=1\textwidth]{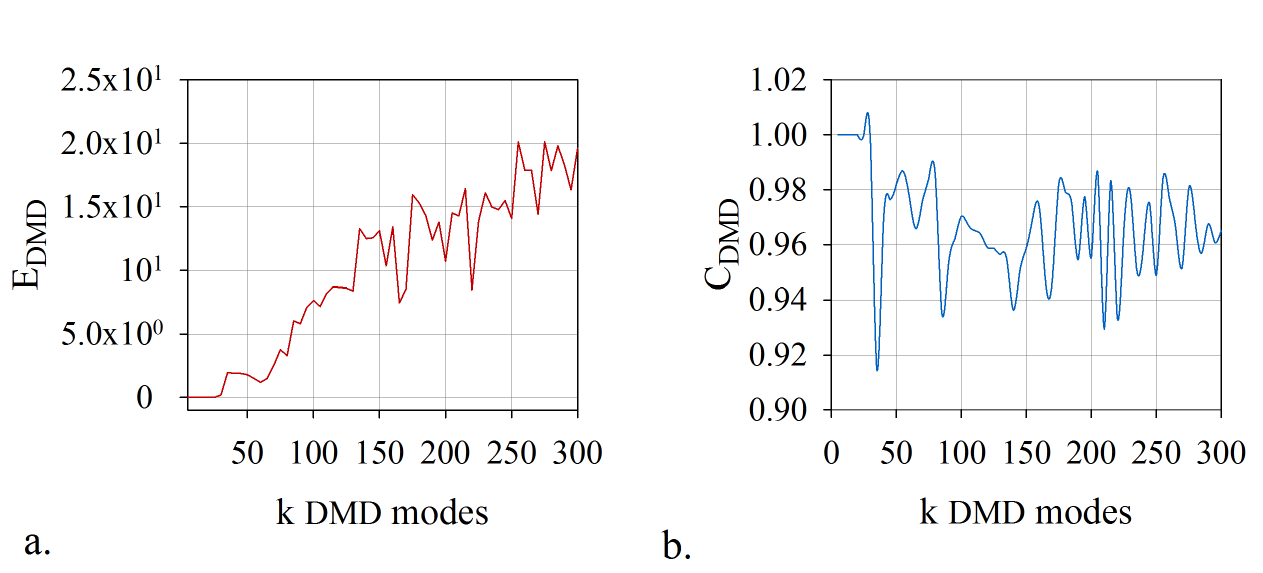}
\caption{Case of $Re=10^2$: a) The relative error computed as a function of retained number of dynamic modes,  b) The correlation coefficient computed as a function of retained number of dynamic modes. $N_{DMD}=15$ leading modes have been selected.}\label{fig4er}
\end{figure}

\begin{figure}[h!]
\centering
\includegraphics[width=1\textwidth]{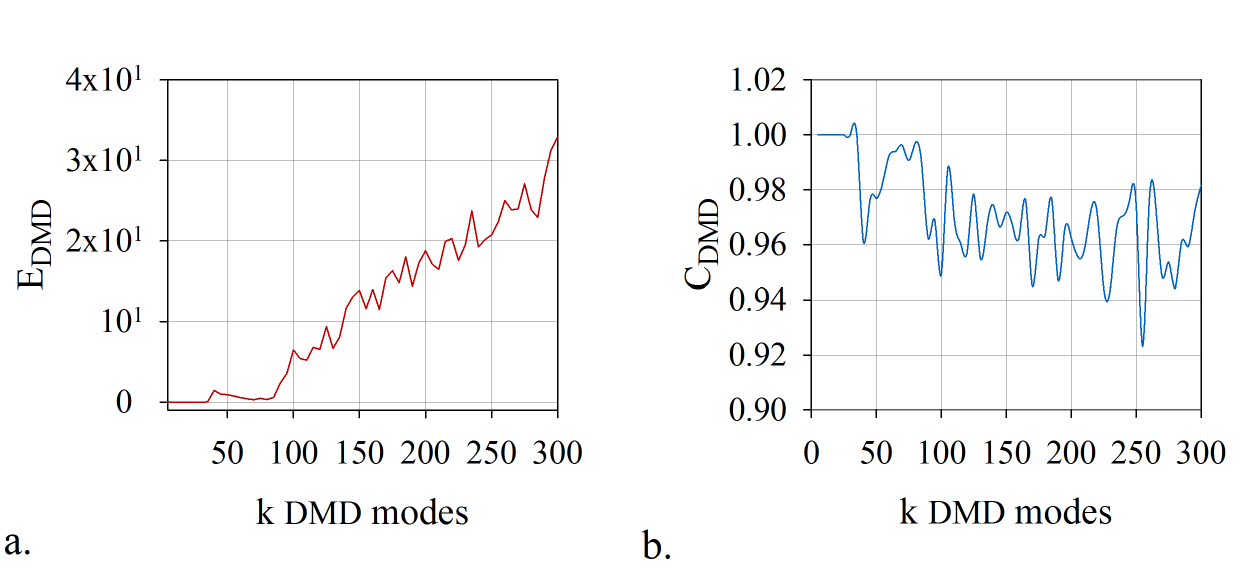}
\caption{Case of $Re=10^3$: a) The relative error computed as a function of retained number of dynamic modes,  b) The correlation coefficient computed as a function of retained number of dynamic modes. $N_{DMD}=20$ leading modes have been selected.}\label{fig5er}
\end{figure}

\begin{figure}[h!]
\centering
\includegraphics[width=1\textwidth]{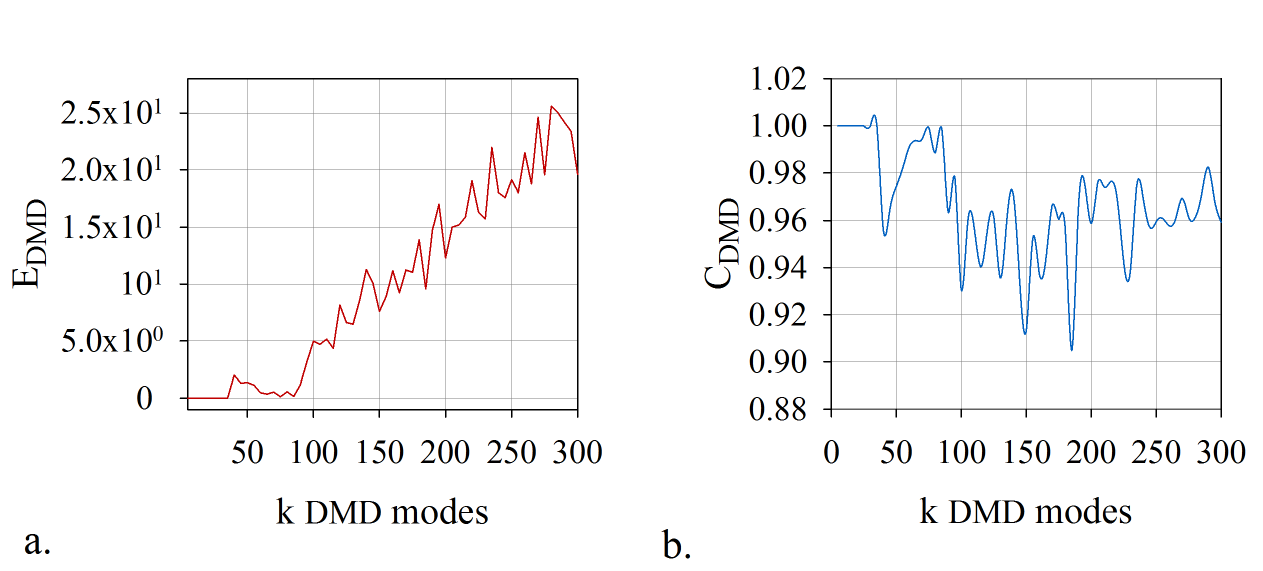}
\caption{Case of $Re=10^4$: a) The relative error computed as a function of retained number of dynamic modes,  b) The correlation coefficient computed as a function of retained number of dynamic modes. $N_{DMD}=20$ leading modes have been selected.}\label{fig6er}
\end{figure}

Figures \ref{fig4er}--\ref{fig6er} present the process of evaluation of DMD model target rank. The error computed as a function of retained number of
dynamic modes and the correlation coefficient are presented, respectively, in the cases $Re=10^2$, $Re=10^3$ and $Re=10^4$.
Table \ref{table1} presents the order of DMD subspace obtained in the three test cases, next to the error defined by Eq. (\ref{eror}) and correlation
coefficient defined by Eq. (\ref{corelation}).

\begin{table}[h!]
\caption{Comparison of the numerical results returned by the randomized dynamic mode decomposition algorithm. $300$ data snapshots have been processed.}\label{table1}
\begin{center}
\begin{tabular}{cccc}
\hline\noalign{\smallskip}
\begin{small}  Test case  \end{small}&\begin{small}  Model rank  \end{small}& \begin{small} Error \end{small} & \begin{small} Correlation coefficient \end{small}\\
\noalign{\smallskip}\hline\noalign{\smallskip}
 $Re=10^2$ & $ N_{DMD}=15$ &  ${E_{DMD}} = {\rm{3}}{\rm{.1984}} \times {\rm{1}}{{\rm{0}}^{ - 7}}$  & ${C_{DMD}} = {\rm{1}}{\rm{.0000}}$\\
 $Re=10^3$ & $ N_{DMD}=20$ & ${E_{DMD}} = {\rm{4}}{\rm{.4247}} \times {\rm{1}}{{\rm{0}}^{ - 7}}$ & ${C_{DMD}} = {\rm{1}}{\rm{.0000}}$\\
$Re=10^4$ &$ N_{DMD}=20$ & ${E_{DMD}} = {\rm{5}}{\rm{.4416}} \times {\rm{1}}{{\rm{0}}^{ - 8}}$ & ${C_{DMD}} = {\rm{1}}{\rm{.0000}}$\\
\hline\noalign{\smallskip}
\end{tabular}
\end{center}
\end{table}

\begin{figure}[h!]
\centering
\includegraphics[width=1\textwidth]{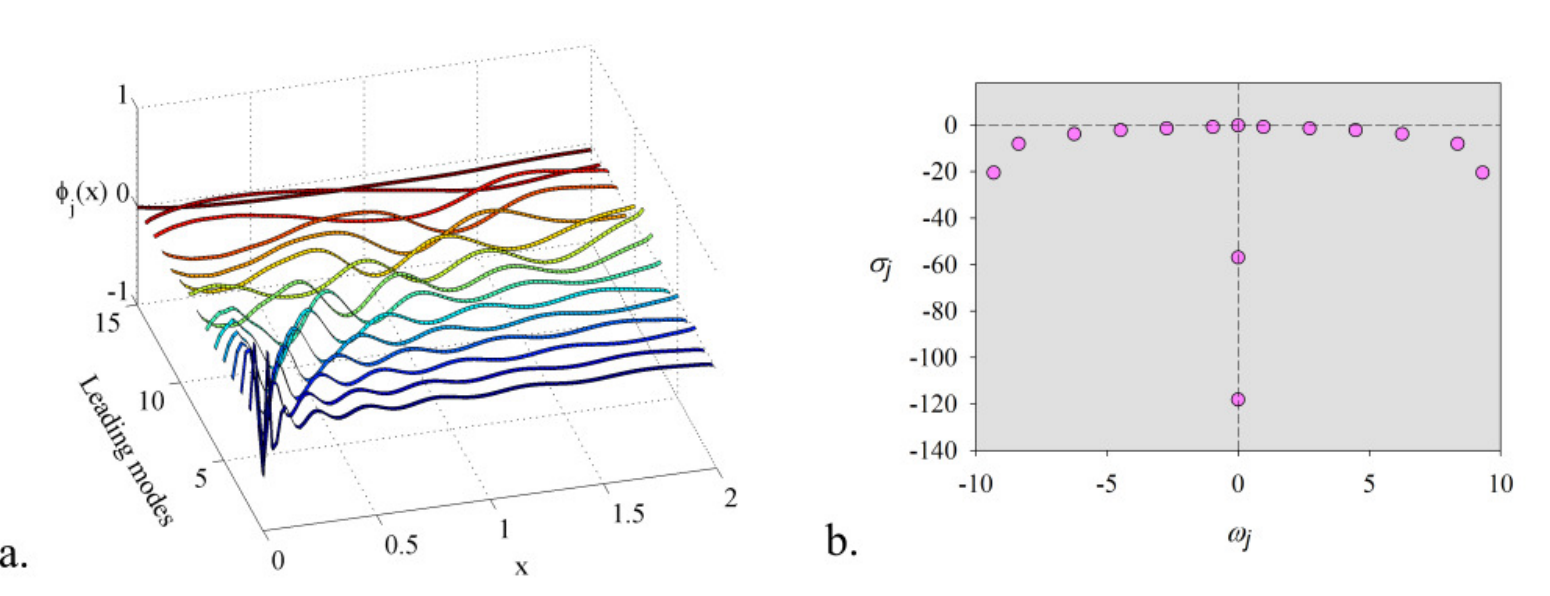}
\caption{Case of $Re=10^2$: a) The DMD leading modes,  b) Growth rates and associated frequencies $\left( {\sigma ,\omega } \right)$ of the eigenvectors of the Koopman matrix $\mathcal{S}$.}\label{fig7mod}
\end{figure}

\begin{figure}[h!]
\centering
\includegraphics[width=1\textwidth]{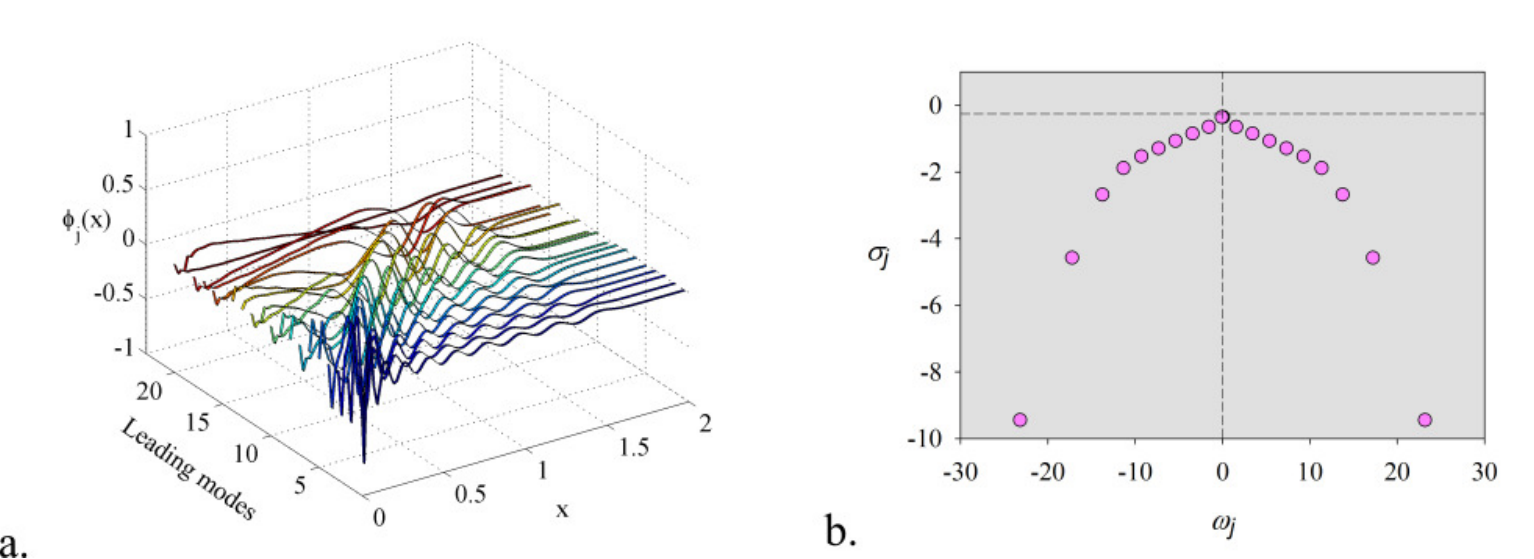}
\caption{Case of $Re=10^3$: a) The DMD leading modes,  b) Growth rates and associated frequencies $\left( {\sigma ,\omega } \right)$ of the eigenvectors of the Koopman matrix $\mathcal{S}$.}\label{fig8mod}
\end{figure}

\begin{figure}[h!]
\centering
\includegraphics[width=1\textwidth]{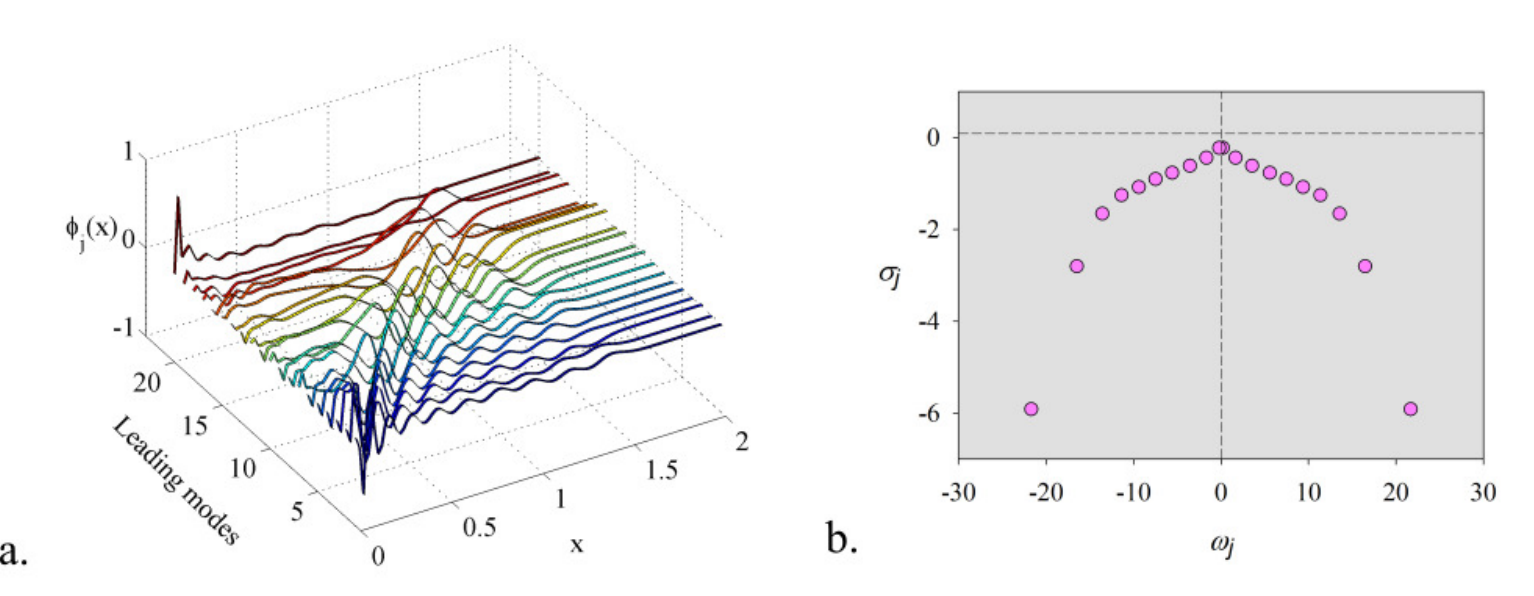}
\caption{Case of $Re=10^4$: a) The DMD leading modes,  b) Growth rates and associated frequencies $\left( {\sigma ,\omega } \right)$ of the eigenvectors of the Koopman matrix $\mathcal{S}$.}\label{fig9mod}
\end{figure}

The algorithm introduced in this paper  confers the best correlation coefficient to the DMD model (see Table \ref{table1}), thus we have identified a digital twin data model. The DMD leading modes are illustrated in Figures \ref{fig7mod}-\ref{fig9mod}, next to the representation of the modal growth rates and the
associated frequencies of the eigenvectors of the Koopman matrix $\mathcal{S}$ for the three test cases, respectively.

The coefficients ${\widehat a_j}\left( t \right)$, $j = 1,...,{N_{DMD}}$ of the reduced order model (\ref{rommodel}) have been estimated for the
entire time window by considering the DMD computed coefficients as inputs of the NLARX model (\ref{nlarx}). The numbers of the input terms, output terms and the value of delay are presented in Tables \ref{table2}-\ref{table4}, for the three test cases, respectively.

The following metrics have been used to perform a qualitative analysis of the digital twin data models (DTMs):
\begin{equation}\label{rommetrix1}
{E_{DTM}} = {\left\langle {{{\left\| {u\left( {x,t} \right) - u_{DTM}^{ROM}\left( {x,t} \right)} \right\|}_2}} \right\rangle _T},
\end{equation}
\begin{equation}\label{rommetrix2}
{C_{DTM}} = \frac{{{{\left\langle {{{\left\| {u\left( {x,t} \right) \cdot u_{DTM}^{ROM}\left( {x,t} \right)} \right\|}_2}} \right\rangle }_T}^2}}{{{{\left\langle {{{\left\| {u{{\left( {x,t} \right)}^H} \cdot u\left( {x,t} \right)} \right\|}_2}} \right\rangle }_T}{{\left\langle {{{\left\| {u_{DTM}^{ROM}{{\left( {x,t} \right)}^H} \cdot u_{DTM}^{ROM}\left( {x,t} \right)} \right\|}_2}} \right\rangle }_T}}},
\end{equation}
where ${E_{DTM}}$ measures the error of the digital twin data model, ${C_{DTM}}$ is the correlation coefficient of the digital twin data model, $u\left( t,x \right)$
means the numerical data, ${u_{DTM}^{ROM}\left( {t,x} \right)}$  represent the computed solution by means of the DMD-ROM model, $\left(  \cdot \right)$ represents the Hermitian inner product, $H$ denotes the conjugate transpose and ${\left\langle  \cdot \right\rangle _T}$ is the norm defined
by Eq. (\ref{Tnorm}).
\begin{table}[h!]
\caption{Initial data for NLARX estimator of temporal coefficients, case of $Re=10^2$, $N_{DMD}=15$.}\label{table2}
\begin{center}
\begin{tabular}{ccc}
\hline\noalign{\smallskip}
\begin{small}  Index  \end{small}&\begin{small}  Outputs, inputs, delay  \end{small}& \begin{small} DTM Error and Correlation coefficient \end{small}\\
\noalign{\smallskip}\hline\noalign{\smallskip}
 $j=1,3-12,14$ & $ n_a=1,n_b=2,n_k=1$ &  ${E_{DTM}} = 8.0806 \times {10^{ - 4}}$   \\
 $j=2$ & $ n_a=2,n_b=2,n_k=2$ &  ${C_{DTM}} = {\rm{1}}{\rm{.0000}}$  \\
 $j=13,15$ & $ n_a=1,n_b=1,n_k=1$ &    \\
\hline\noalign{\smallskip}
\end{tabular}
\end{center}
\end{table}
\begin{table}[h!]
\caption{Initial data for NLARX estimator of temporal coefficients, case of $Re=10^3$, $N_{DMD}=20$.}\label{table3}
\begin{center}
\begin{tabular}{ccc}
\hline\noalign{\smallskip}
\begin{small}  Index  \end{small}&\begin{small}  Outputs, inputs, delay  \end{small}& \begin{small} DTM Error and Correlation coefficient \end{small}\\
\noalign{\smallskip}\hline\noalign{\smallskip}
 $j=1,4,10-12$ & $ n_a=2,n_b=1,n_k=1$ &  ${E_{DTM}} = 1.3080 \times {10^{ - 4}}$  \\
 $j=2,3,13$ & $ n_a=1,n_b=1,n_k=2$ &  ${C_{DTM}} = {\rm{1}}{\rm{.0000}}$  \\
 $j=5-9$ & $ n_a=1,n_b=1,n_k=5$ &    \\
 $j=14-19$ & $ n_a=2,n_b=1,n_k=2$ &    \\
 $j=20$ & $ n_a=1,n_b=1,n_k=1$ &    \\
\hline\noalign{\smallskip}
\end{tabular}
\end{center}
\end{table}
\begin{table}[h!]
\caption{Initial data for NLARX estimator of temporal coefficients, case of $Re=10^4$, $N_{DMD}=20$.}\label{table4}
\begin{center}
\begin{tabular}{ccc}
\hline\noalign{\smallskip}
\begin{small}  Index  \end{small}&\begin{small}  Outputs, inputs, delay  \end{small}& \begin{small} DTM Error and Correlation coefficient \end{small}\\
\noalign{\smallskip}\hline\noalign{\smallskip}
 $j=1,3,4,7$ & $ n_a=1,n_b=3,n_k=2$ &  ${E_{DTM}} = 1.4000 \times {10^{ - 4}}$  \\
 $j=2,5,6,8,16-20$ & $ n_a=1,n_b=2,n_k=2$ &  ${C_{DTM}} = {\rm{1}}{\rm{.0000}}$  \\
 $j=9-15$ & $ n_a=1,n_b=1,n_k=1$ &    \\
\hline\noalign{\smallskip}
\end{tabular}
\end{center}
\end{table}

\begin{figure}[h!]
\centering
\includegraphics[width=1\textwidth]{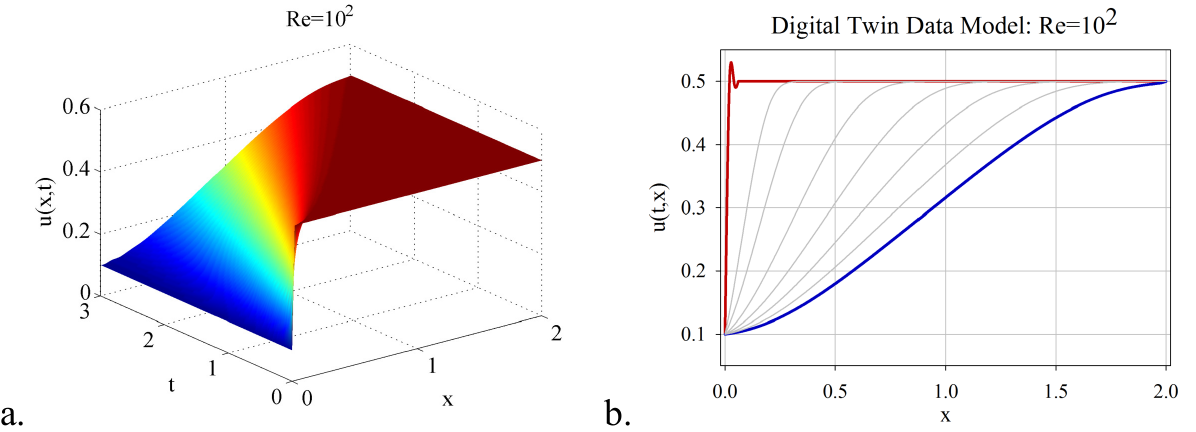}
\caption{Solution of the digital twin data model in the case of experiment $Re=10^2$: a) 3D view; b) Projection view of the shock wave. }\label{DTMRe2}
\end{figure}

\begin{figure}[h!]
\centering
\includegraphics[width=1\textwidth]{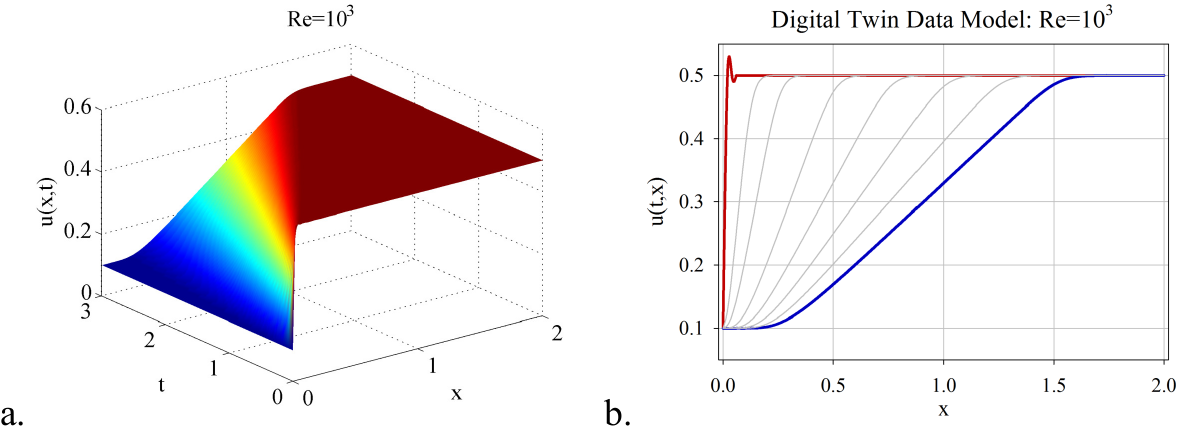}
\caption{Solution of the digital twin data model in the case of experiment $Re=10^3$: a) 3D view; b) Projection view of the shock wave. }\label{DTMRe3}
\end{figure}

\begin{figure}[h!]
\centering
\includegraphics[width=1\textwidth]{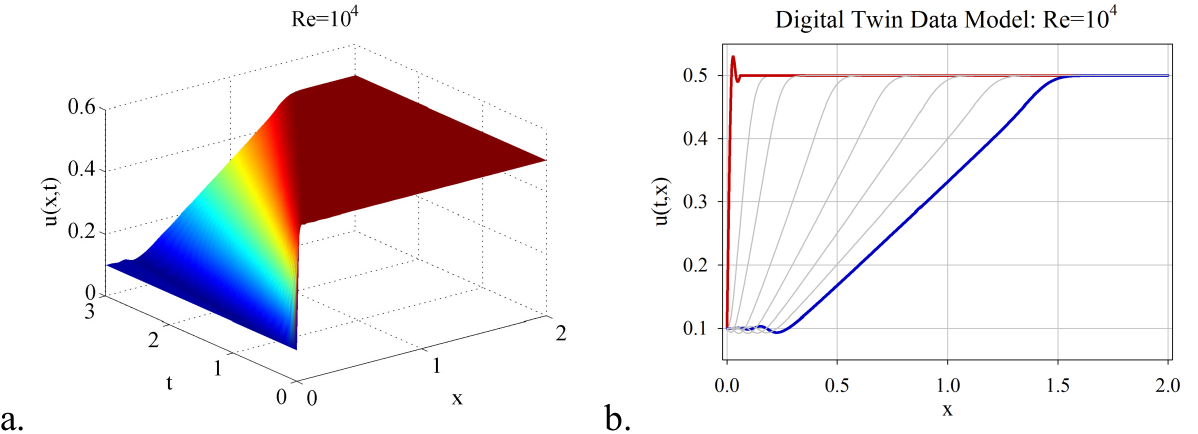}
\caption{Solution of the digital twin data model in the case of experiment $Re=10^4$: a) 3D view; b) Projection view of the shock wave. }\label{DTMRe4}
\end{figure}

Solution of the digital twin data models are illustrated in Figures \ref{DTMRe2}-\ref{DTMRe4}, for the three test cases,
respectively. The very good correlation coefficients and the low value of errors presented in Tables \ref{table2}-\ref{table4} confirm the
computational efficiency of the digital twin data models.

\begin{figure}[h!]
\centering
\includegraphics[width=1\textwidth]{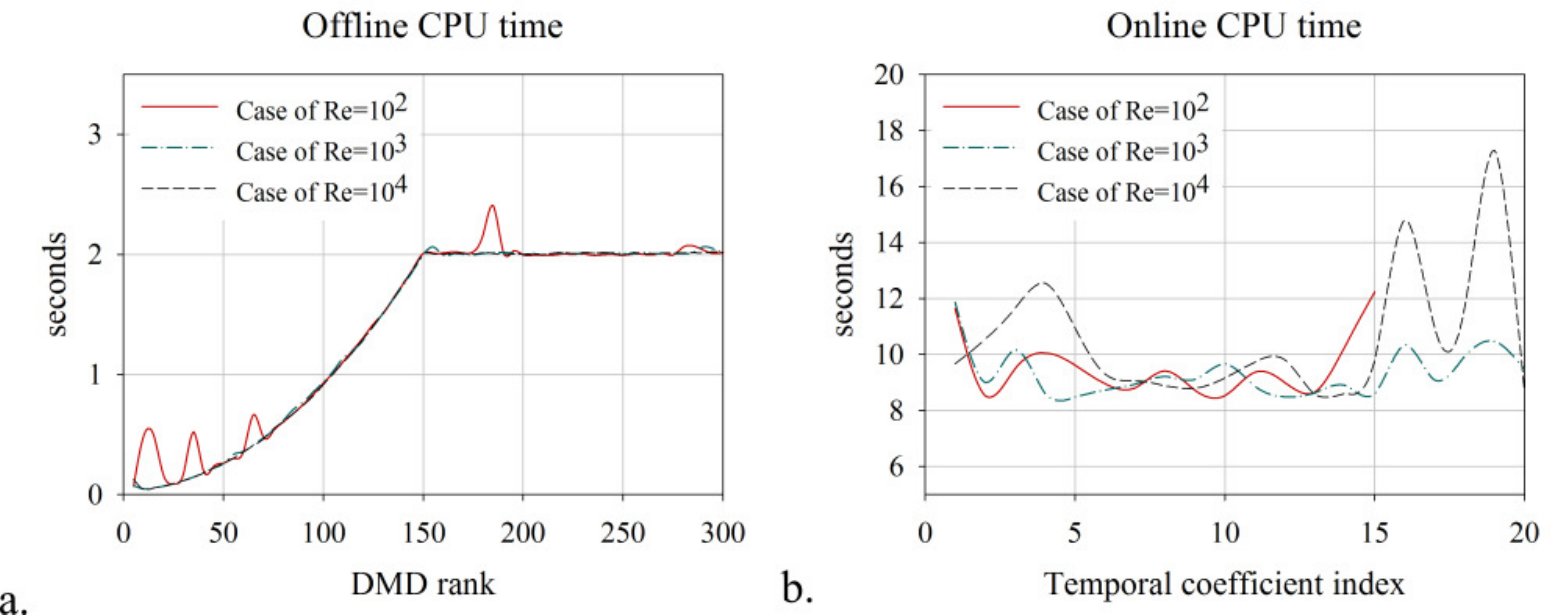}
\caption{The CPU time required in the offline-online stage, for the three test cases.}\label{fig13cpu}
\end{figure}

The CPU time required in the offline-online stage is presented in Figure \ref{fig13cpu}, for the three test cases.
The required offline CPU time does not exceed two seconds and does not present large variations depending on the case study. The online CPU time fall between $8$ and $17$ seconds, depending on the index of the temporal coefficient which is estimated along the entire time window. It is obvious that the NLARX estimator requires more time to estimate the temporal behaviour in the case of very high Reynolds number.

\begin{figure}[h!]
\centering
\includegraphics[width=0.9\textwidth]{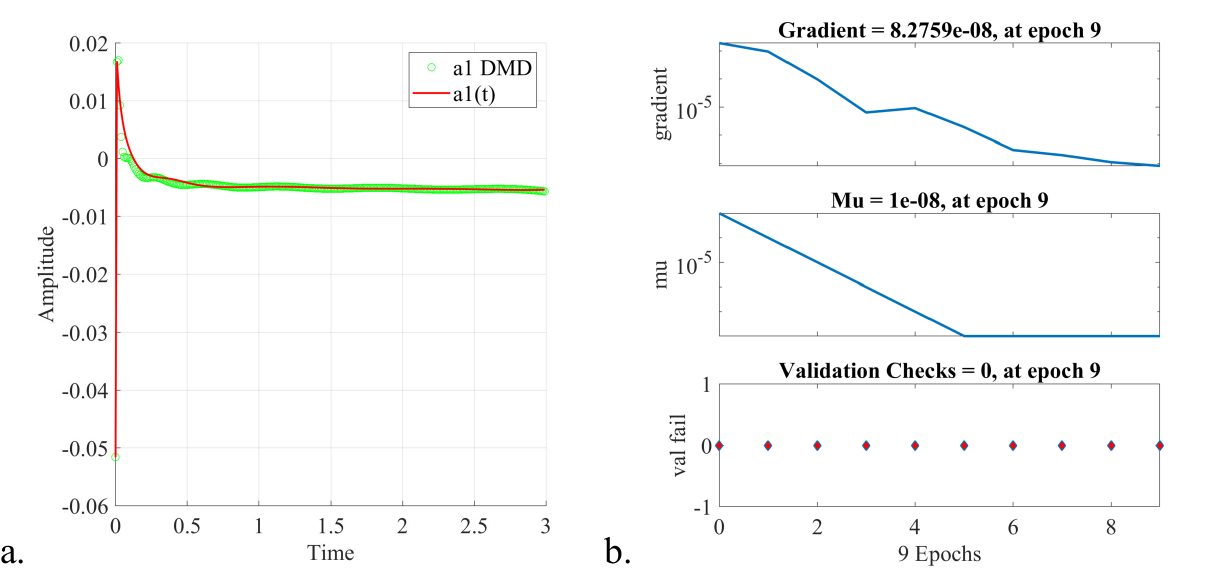}
\caption{The validation for the first temporal coefficient, as simulated response of the optimal NLARX estimator, in the
case of experiment ${\mathop{\rm Re}\nolimits}  = {10^2}$.}\label{coef1Re2}
\end{figure}

\begin{figure}[h!]
\centering
\includegraphics[width=0.9\textwidth]{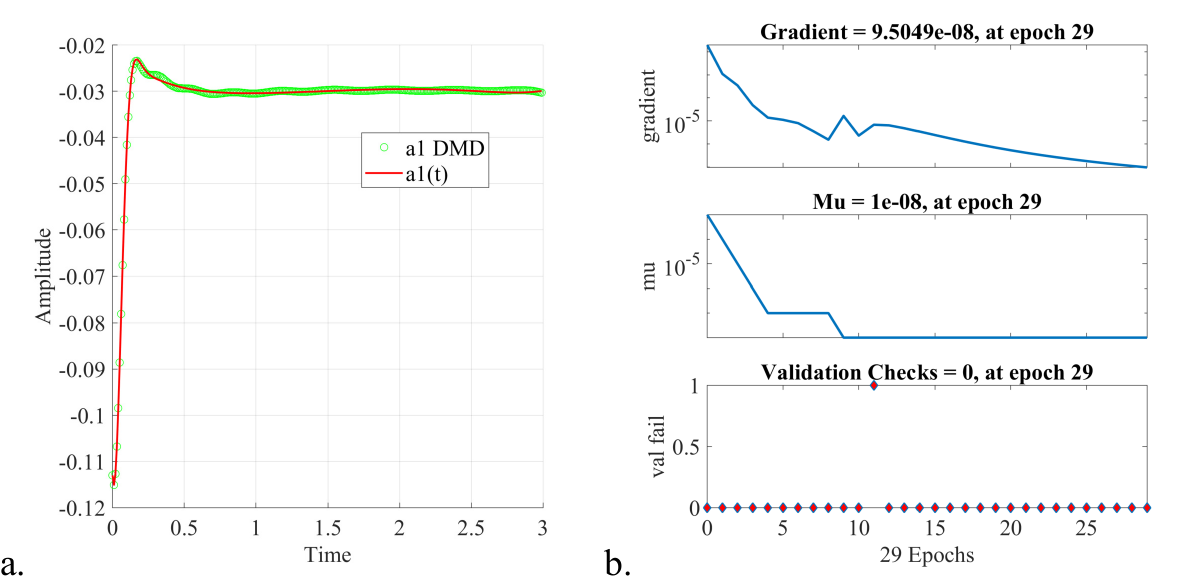}
\caption{The validation for the first temporal coefficient, as simulated response of the optimal NLARX estimator, in the
case of experiment ${\mathop{\rm Re}\nolimits}  = {10^3}$.}\label{coef1Re3}
\end{figure}

\begin{figure}[h!]
\centering
\includegraphics[width=0.9\textwidth]{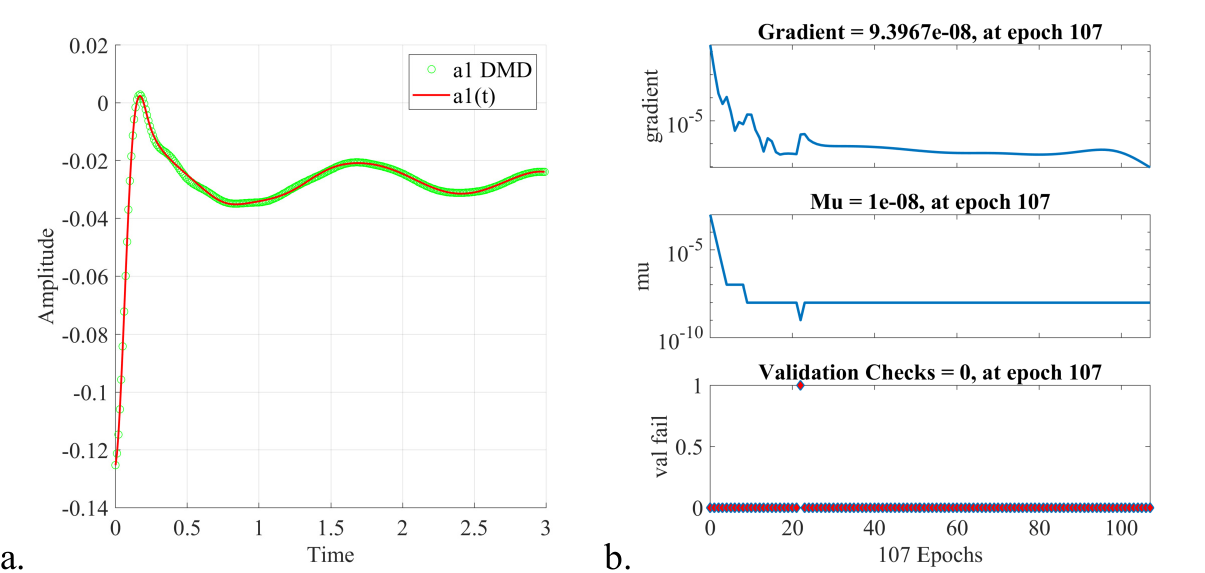}
\caption{The validation for the first temporal coefficient, as simulated response of the optimal NLARX estimator, in the
case of experiment ${\mathop{\rm Re}\nolimits}  = {10^4}$.}\label{coef1Re4}
\end{figure}

Figures \ref{coef1Re2}-\ref{coef1Re4} illustrate the validation for the first
 temporal coefficient, as simulated response of the optimal NLARX estimator, in the
case of the three experiments, respectively.

\section{Conclusions}

The present investigation has focused on the identification of high-fidelity digital twin data models from numerical code outputs by non-intrusive techniques (i.e. not requiring Galerkin projection of the governing equations onto the reduced modes basis). In this paper we define the concept of digital twin data model (DTM) as a model of reduced complexity that has the main feature to mirror the original process behavior.

 We developed an algorithm that utilizes a variant of adaptive randomized dynamic mode decomposition introduced in
\cite{randomized2017} to obtain a reduced basis in the offline stage, combined with a fast digital twin identification using neural network based
nonlinear autoregressive estimators in the online stage. To overcome the inconveniences of developing and implementing a mode selection criterion associated with dynamic mode decomposition,  we developed a  technique based on randomized dynamic mode decomposition as a fast and accurate option in model order reduction. The rank of the ROM is given as the unique solution of an optimization problem whose constraint consists in the smallest error of DTM. Solving the
     optimization problem (\ref{optimprob}) using a  hybrid simulated annealing \cite{SA} we gain a fast and accurate randomized DMD algorithm, with a low rank for the ROMs.

 The DTMs have been investigated in the numerical simulation of three shock wave phenomena with increasing complexity, with Reynolds number varying from $10^2$  to $10^4$.

 We showed that the significant advantage of DTM is to map the dynamics with high accuracy and reduced costs in CPU time and hardware, even to timescales difficult to explore because of the rapidly changing dynamics over time.

The procedure of online estimation of the DTM temporal coefficients by employing neural network based nonlinear autoregressive estimators
    leads to a fast and accurate identification of the digital twin data models, as seen in Tables \ref{table2}-\ref{table4}. We investigated the computational
    efficiency of the proposed algorithm and we provided a qualitative analysis of the DTM in the three experiments investigated.

\end{document}